\documentclass{article}

\usepackage{microtype}
\usepackage{graphicx}
\usepackage{subcaption}
\usepackage{booktabs} 

\usepackage{xspace}
\newcommand{\project}{$\text{3DGS}^2$-TR\xspace}

\usepackage{listings}
\usepackage{xcolor}

\lstdefinestyle{cppstyle}{
  language=C++,
  backgroundcolor=\color{gray!10},
  basicstyle=\ttfamily\footnotesize,
  keywordstyle=\color{blue!90}\bfseries,
  commentstyle=\color{gray!70}\itshape,
  stringstyle=\color{orange!90!black},
  numbers=left,
  numberstyle=\tiny\color{gray},
  stepnumber=1,
  numbersep=6pt,
  frame=single,
  rulecolor=\color{gray!30},
  breaklines=true,
  tabsize=2,
  showstringspaces=false,
  captionpos=b
}
\usepackage{multirow}
\usepackage{threeparttable}
\usepackage{makecell}

\usepackage{graphicx}  
\usepackage{stfloats}     
\usepackage{balance}

\usepackage{hyperref}

\usepackage[preprint]{icml2026}

\usepackage{amsmath}
\usepackage{amssymb}
\usepackage{mathtools}
\usepackage{amsthm}

\usepackage[capitalize,noabbrev]{cleveref}

\theoremstyle{plain}

\theoremstyle{definition}

\theoremstyle{remark}

\usepackage[textsize=tiny]{todonotes}

\icmltitlerunning{Scalable Second-Order Optimizer with Hellinger Distance Trust-Region for Large-Scale 3D Gaussian Splatting}

\begin{document}

\twocolumn[
  \icmltitle{\project: Scalable Second-Order Trust-Region Method \\ for 3D Gaussian Splatting}



  \icmlsetsymbol{equal}{*}

  \begin{icmlauthorlist}
    \icmlauthor{Roger Hsiao}{UCB}
    \icmlauthor{Yuchen Fang}{UCB}
    \icmlauthor{Xiangru Huang}{UCB}
    \icmlauthor{Ruilong Li}{Nvidia}
    \icmlauthor{Hesam Rabeti}{Nvidia}
    \icmlauthor{Zan Gojcic}{Nvidia}
    \icmlauthor{Javad Lavaei}{UCB}
    \icmlauthor{James Demmel}{UCB}
    \icmlauthor{Sophia Shao}{UCB}
  \end{icmlauthorlist}

  \icmlaffiliation{UCB}{University of California, Berkeley}

    \icmlaffiliation{Nvidia}{NVIDIA}
  \icmlcorrespondingauthor{Roger Hsiao}{roger\_hsiao@berkeley.edu}
  
  \icmlkeywords{Machine Learning, ICML}

  \vskip 0.3in
]



\printAffiliationsAndNotice{}  

\begin{abstract}
    We propose \project, a second-order optimizer for accelerating the scene training problem in 3D Gaussian Splatting (3DGS). 
    Unlike existing second-order approaches that rely on explicit or dense curvature representations, such as 3DGS-LM \cite{hollein20253dgs}  or 3DGS$^2$\cite{lan20253dgs2},
    our method approximates curvature using only the diagonal of the Hessian matrix, estimated efficiently via Hutchinson’s method. 
    Our approach is fully matrix-free and has the same complexity as ADAM \cite{kingma2014adam}, $O(n)$ in both computation and memory costs.
    To ensure stable optimization in the presence of strong nonlinearity in the 3DGS rasterization process, we introduce a parameter-wise trust-region technique based on the squared Hellinger distance, regularizing updates to Gaussian parameters.
    Under identical parameter initialization and without densification, \project is able to achieve better reconstruction quality on standard datasets, using 50\% fewer training iterations compared to ADAM, while incurring less than 1GB of peak GPU memory overhead (17\% more than ADAM and 85\% less than 3DGS-LM),
    enabling scalability to very large scenes and potentially to distributed training settings.
\end{abstract}

\section{Introduction}
\label{sec:intro}

Recent advances in radiance field representations have revolutionized 3D content creation, enabling high-fidelity, photorealistic scene reconstruction from sparse input views. 
Neural Radiance Fields (NeRF) \cite{mildenhall2021nerf} pioneered coordinate-based neural scene representations, achieving remarkable rendering quality but suffering from slow training and inference times. 
To address these limitations, 3D Gaussian Splatting (3DGS) \cite{kerbl20233d} proposed an explicit and efficient scene representation that models geometry and appearance using anisotropic 3D Gaussians optimized through differentiable rasterization. This formulation preserves view-dependent effects while enabling real-time rendering, rapidly establishing 3DGS as the new standard for real-time novel view synthesis—the task of rendering photorealistic images from previously unseen camera viewpoints given a limited set of input images. This capability is central to applications in virtual reality, augmented reality, robotics, and 3D content creation.

\begin{figure}[t]
    \centering
\includegraphics[width=0.98\linewidth]{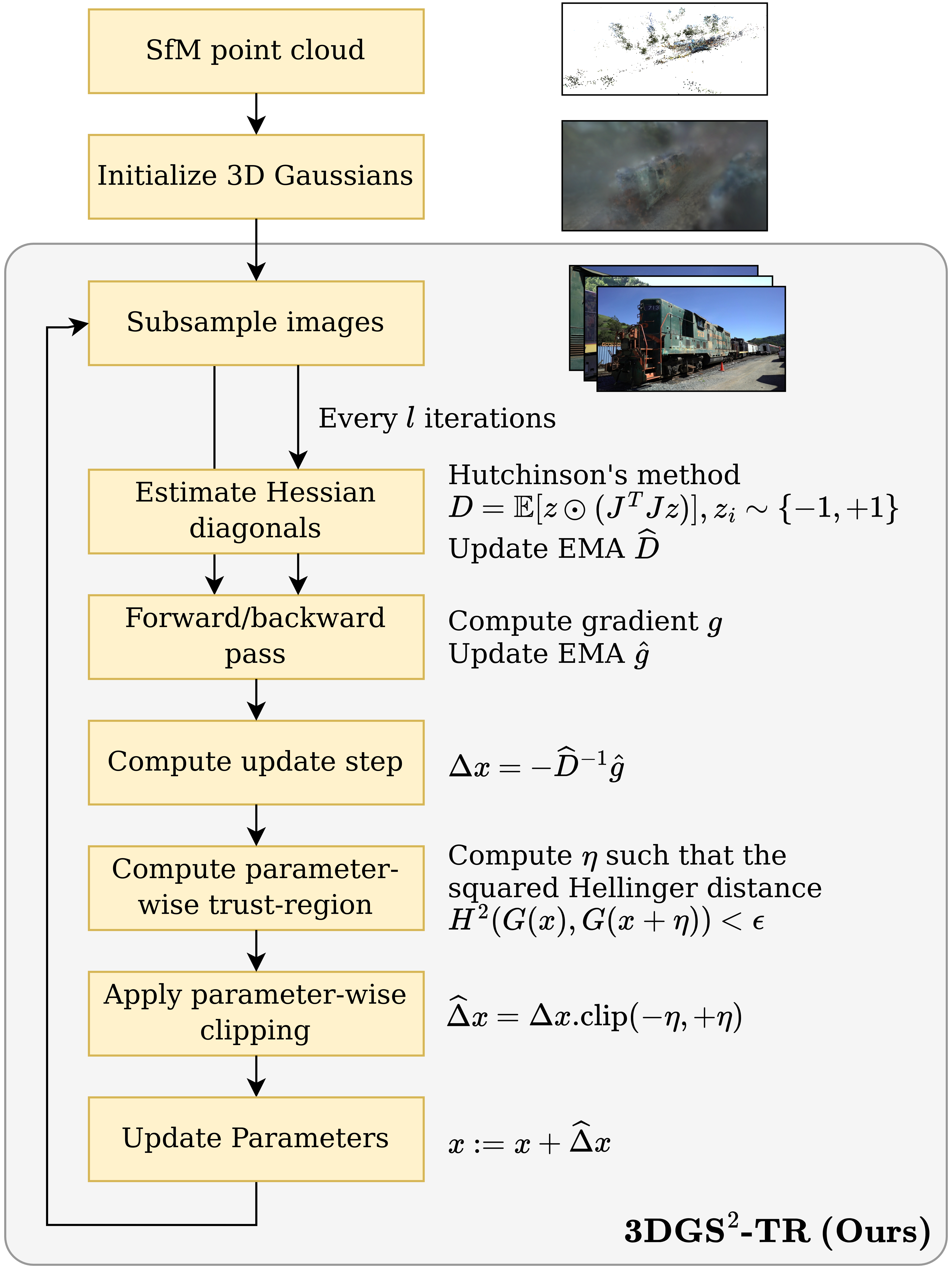}
    \caption{Overview of our proposed method.}
    \label{fig:overview1}
\end{figure}

Building on the original 3DGS framework's demonstration that anisotropic Gaussians coupled with GPU-friendly rasterization can achieve both high-quality synthesis and real-time performance \cite{kerbl20233d}, a number of subsequent works have proposed strategies to further improve rasterization efficiency. Recent efforts focus on accelerating forward rasterization and inference through optimized GPU kernels, memory access patterns, and tile-based rendering strategies, enabling efficient large-scale deployment and real-time visualization \citep{durvasula2023distwar,feng2025flashgs,mallick2024taming,ye2025gsplat,papantonakis2024reducing,zhao2024scaling}.

Despite these advances in rasterization and scene representation, scene training in 3DGS remains a bottleneck in the pipeline. 
Although significantly faster than NeRF, 3DGS still requires training from scratch for each new scene, taking 20-40 minutes on commercial GPUs for commonly used datasets \cite{kerbl20233d}. 
This limits the applicability of 3DGS in large-scale or real-time exploration scenarios.

Several lines of research address this training bottleneck through orthogonal approaches. One direction targets the backward pass, which accumulates gradients into Gaussian splats with poor GPU utilization due to resource contention when many pixels share the same Gaussian primitives. Recent work has identified this as a dominant bottleneck and proposed alternative accumulation schemes or kernel redesigns to mitigate atomic contention \citep{durvasula2023distwar,ye2025gsplat}. Another approach reduces training time through more compact scene representations. Related efforts improve the densification process or introduce pruning strategies to reduce the total number of Gaussian primitives while preserving reconstruction quality, thereby indirectly accelerating training speed \citep{kheradmand20243d,fang2024mini,rota2024revising,ali2024trimming,zhang2024lp,hanson2025pup}.

Our work aligns most closely with a third direction: improving the optimizer itself. The standard optimizer for 3DGS is ADAM \citep{kingma2014adam}, a first-order gradient-descent method widely used in deep learning. While first-order optimizers are easy to implement and scale well, they suffer from slow convergence in highly non-convex and ill-conditioned parameter spaces \citep{sutton1986two,dauphin2014identifying,bottou2018optimization}. The optimization landscape of 3DGS is particularly challenging due to the strong coupling between geometry parameters (position, rotation, scale) and appearance parameters (opacity, color), leading to inefficient convergence and excessive training iterations.
Additionally, first-order methods require meticulous tuning of learning rates for each parameter \citep{sutton1986two,schaul13no,dauphin2014identifying,bottou2018optimization}.

3DGS occupies a unique position in machine learning: 
the model quality scales with the number of parameters, yet each parameter remains highly interpretable as a component of an unnormalized 3D Gaussian distribution. 
This has motivated prior work to explore second-order optimization algorithms—such as Newton's method, Gauss-Newton, or Levenberg-Marquardt—to achieve superlinear convergence \citep{hollein20253dgs,lan20253dgs2,pehlivan2025second}.
However, the 3DGS rasterization function poses fundamental challenges for second-order methods. First, it is highly nonlinear due to the sequential rendering of Gaussian splats at each pixel, where the transmittance seen by each splat depends on the opacities of all preceding splats in depth order. 
Second, the loss function is only piecewise continuous since depth-based sorting of Gaussian splats causes render order to change discontinuously with position parameters. 
These discontinuities become more frequent in regions with dense clusters of splats, which commonly occur towards the end of training.
Moreover, existing implementations of second-order methods incur high memory overhead from storing curvature information and suffer from expensive per-iteration costs due to matrix operations, offsetting any convergence speedup and making them unsuitable for production use with large scenes. 

In practice, second-order optimizers for 3DGS have yet to outperform first-order methods in either reconstruction quality or convergence speed, leading to limited adoption. However, recent work in deep learning, such as Sophia \cite{liu2023sophia}, introduces a lightweight, second-order approach that utilizes a diagonal Hessian estimate to account for loss surface curvature. 
By adapting step sizes to sharp or flat regions with minimal computational overhead, Sophia achieves faster convergence and superior stability in large-scale non-convex tasks such as large language model training.
Nevertheless, the performance of Sophia and its variants remains largely unexplored in computer vision applications.

Based on these observations, we propose the following three principles to accelerate scene training in 3DGS. 
\begin{enumerate}
    \item \textit{Cheap per-iteration computation.} Since 3DGS rasterization is highly nonlinear with frequent discontinuities, we favor many small, inexpensive steps over a few large, expensive ones that would likely violate local linear or quadratic approximations. Given that the backward pass is the primary training bottleneck, our strategy introduces modest computational overhead in pre- or post-processing the update step to accelerate convergence and reduce the total number of backward passes required.
    \item \textit{Parameter-linear memory scaling.} Persistent storage across iterations should scale with the number of Gaussian parameters, not with the number of pixels, which often exceed the parameter count by orders of magnitude for large scenes with high-resolution training images.
    \item \textit{Trust region constraints.} Update step sizes should be bounded by a trust region that limits the impact of nonlinear interactions between Gaussian splats. Specifically, we bound the change in transmittance experienced by each splat before and after the update, ensuring steps remain within the region of validity for local approximation.
\end{enumerate}

Guided by these principles, we present \project, a second-order optimizer for accelerated 3DGS scene training. 
We summarize our main contributions as follows:
\begin{enumerate}
    \item We apply a second-order update rule that uses Hessian diagonals to 3DGS training, which has the same $O(n)$ complexity as ADAM in both computation and memory. 
    \item We propose an effective trust region for Gaussian parameter updates that bounds the squared Hellinger distance of each Gaussian splat before and after optimization steps, providing a principled constraint on geometric changes and requiring only a single tunable hyperparameter.
\end{enumerate}

\section{Related Work}

\subsection{Novel View Synthesis}
Novel view synthesis aims to generate photorealistic images from arbitrary viewpoints given a set of input images. Neural Radiance Fields (NeRF) \cite{mildenhall2021nerf} pioneered implicit scene representations using multilayer perceptrons to encode volumetric density and view-dependent appearance, achieving photorealistic results through volumetric ray marching but suffering from slow training and inference.
In contrast, 3D Gaussian Splatting (3DGS) \cite{kerbl20233d} models scenes explicitly as collections of anisotropic 3D Gaussians, enabling real-time rendering through efficient differentiable rasterization while maintaining high visual fidelity. This has established 3DGS as the state-of-the-art for real-time novel view synthesis, inspiring extensive research on improving rendering quality \cite{lu2024scaffold,yu2024mip}, scaling to larger environments \cite{kerbl2024hierarchical,song2024city}, and enhancing geometric accuracy.

\subsection{Accelerating 3DGS Training and Rendering}
While 3DGS achieves real-time rendering, training remains a bottleneck for large-scale scenes. Recent work addresses this through complementary strategies.

\textbf{Compact representations.} Several methods aim to reduce the number of Gaussian primitives while preserving quality. EAGLES \cite{girish2024eagles} uses quantized embeddings and coarse-to-fine training. LightGaussian \cite{fan2024lightgaussian} prunes low-contribution Gaussians and compresses spherical harmonics. C3DGS \cite{lee2024compact} learns binary masks to remove redundant primitives, while Speedy-Splat \cite{hanson2025speedy} introduces tight bounding boxes and dual pruning strategies.

\textbf{Rasterization optimizations.} AdR-Gaussian \cite{wang2024adr} culls low-opacity Gaussian-tile pairs and balances workloads. FlashGS \cite{feng2025flashgs} and gsplat \cite{ye2025gsplat} optimize CUDA kernels and memory access patterns. Additional improvements include modified densification heuristics \cite{fang2024mini,kheradmand20243d} that reduce Gaussian proliferation.

\subsection{Second-Order Optimizers}
Recent research in large-scale machine learning explores replacing ADAM \citep{kingma2014adam} with second-order methods to accelerate convergence. Classical second-order methods—Newton's method, Gauss-Newton, and Levenberg-Marquardt—achieve superlinear convergence via Hessian curvature information but face $O(n^2)$ memory and $O(n^3)$ computational costs. Lightweight alternatives such as AdaHessian \citep{yao2021adahessian} and Sophia \citep{liu2023sophia}, instead use diagonal Hessian approximations for efficient large-scale optimization, avoiding the storage and computation overhead while gaining improvement in performance. 

Applications of second-order optimizers have been explored in 3DGS as well. For example, 3DGS-LM \cite{hollein20253dgs} integrates the Levenberg-Marquardt method, while 
3DGS2 \cite{lan20253dgs2} partitions parameters and solves small Newton systems in sequence. Recently, \cite{pehlivan2025second} adopted a matrix-free design, applying the preconditioned conjugate gradient method to solve the system, and introduced a pixel sampling strategy based on residuals.

However, some fundamental challenges have not been adequately addressed: 
3DGS rasterization is highly nonlinear and only piecewise continuous due to depth-based sorting, violating smoothness assumptions of classical second-order methods. 
Moreover, the high parameter and pixel counts make it intractable to materialize full matrices, necessitating careful camera perspective grouping and image subsampling \cite{hollein20253dgs, lan20253dgs2}.  
Consequently, existing approaches only work well on small scenes or when initialized near a sufficiently good local minimum.
In contrast, our work adapts Sophia's lightweight framework to 3DGS with modifications tailored to Gaussian-based representations while maintaining efficiency and broad applicability.
\section{Background and Notations}

\subsection{Review of Gaussian Splatting}
3D Gaussian Splatting (3DGS) \cite{kerbl20233d} represents a scene as a collection of 3D Gaussians splats, each parametrized by its position $\mu$, scale $S$, rotation $R$, opacity $\alpha$, and color $\mathcal{C}$.
We can write each Gaussian as an unnormalized Gaussian distribution
\begin{align}\label{gaussian-primitive}
    G(z) &= \alpha \mathcal{C} \exp(-\frac12 (z - \mu)^T \Sigma^{-1}(z-\mu)) \\
         &= \alpha \mathcal{C} (2\pi)^{\frac32} \det(\Sigma)^{\frac12} \mathcal{N}(z; \mu, \Sigma)
\end{align}
The covariance is split into separate rotation and scaling parameters as $\Sigma = R^TS^TSR$.
$\mathcal{C}$ is the view-dependent color modeled using spherical harmonic coefficients of order 3.
To render an image of size $H\times W$ from a given viewpoint, 
all Gaussians are first projected into 2D Gaussian splats via a tile-based differentiable rasterizer. 
The projected splats are then $\alpha$-blended along each camera ray to obtain the color $C_\omega$ at pixel $\omega$:
\begin{equation*}
    C_\omega = \sum_{i\in \mathcal{K}} \mathcal{C}_i \bar{\alpha}_i T_i, \text{ with }T_i = \prod_{j=1}^{i-1}(1-\bar{\alpha}_j),
\end{equation*}
where $\mathcal{K}$ denotes the set of Gaussian kernels of size $|\mathcal{K}|$, $\mathcal{C}_i$ is the color of the $i$-th splat along the ray, $\bar{\alpha}_i$ is the 2D Gaussian’s evaluated opacity, and $T_i$ represents transmittance. The complete Gaussian parameter vector $x$ concatenates the parameters of all $|\mathcal{K}|$ kernels by groups, (e.g. $x=(x_{position},x_{scaling},x_{rotation},x_{opacity},x_{color})$. 
To fit the Gaussian parameter $x$, 3DGS minimizes the discrepancy between the rendered and ground-truth images. At each pixel $\omega$, the loss function is written as 
\begin{equation}\label{pixel_loss}
    \mathcal{L}_\omega = (1 - \lambda) \mathcal{L}_{\omega, L_1} + \lambda \mathcal{L}_{\omega, D-SSIM}.
\end{equation}
The L1 and SSIM losses can each be further split into three components, one for each RGB channel. 
The total loss is the mean of the loss over all pixels and all channels.

\subsection{Optimization Problem Formulation}
Let $x \in \mathbb{R}^n$ be the parameter vector by concatenating the parameters of all Gaussian kernels.
Let $M$ be the number of training images. 
Each training image emits $6\times H \times W$ loss components with a total of $m$ loss components over all images. 
We use $f_{i}(x)$ to denote the vector loss function for image $i$, where each entry is the square root of a loss component, 
so that the 3DGS scene training problem can be formulated as a nonlinear least-squares problem 
\begin{equation}\label{objective}
    \min_x \; f(x) \coloneqq \frac{1}{2m} \left \lVert \begin{bmatrix}
        f_1(x)^T & \cdots & f_M(x)^T
    \end{bmatrix}^T \right \rVert_2^2.
\end{equation}
The full Jacobian matrix is denoted as $J(x) = \begin{bmatrix}
    J_1(x)^T & \cdots & J_M(x)^T 
\end{bmatrix}^T$ and the pseudo-Hessian matrix is $H(x) = J(x)^TJ(x) = \sum_{i=1}^M J_i(x)^TJ_i(x)$.

\section{Methodology}


\project is a scalable second-order optimization framework for solving the 3DGS training objective~\eqref{objective}. Our method adopts a stochastic Gauss--Newton formulation and incorporates curvature information while remaining memory- and compute-efficient. 
An overview of the proposed pipeline is illustrated in \autoref{fig:overview1}, whereas the full algorithm is summarized in \autoref{alg:3dgs2-tr}. 


\begin{algorithm}[t]
\caption{$\text{3DGS}^2$-TR}
\label{alg:3dgs2-tr}
\begin{algorithmic}[1]
\REQUIRE Initial parameters $x_1\in\mathbb{R}^n$, Hessian diagonals update interval $l \in \mathbb{N}$, 
EMA decay rates $\theta_1,\theta_2 \in (0,1)$, trust-region parameter $\epsilon>0$, 
Hutchinson sample size $\nu\in \mathbb{N}$.
\STATE Set $\widehat{g}_0=0,\widehat{D}_{1-l}=0$.
\FOR{$t = 1$ to $T$}
    \STATE Sample a mini-batch $\mathcal{S}_1$ of images and compute the stochastic gradient $g_t$.
    \STATE Update EMA $ \widehat{g}_k = \theta_1 \widehat{g}_{k-1} + (1-\theta_1) g_k$.
    \IF{$t \mod l =1$}
    \STATE Sample a mini-batch $\mathcal{S}_2$ of images and compute $ D_t= \text{Hutch}(x_t,\mathcal{S}_2,\nu)$.
    \STATE Update EMA $\widehat{D}_t = \theta_2 \widehat{D}_{t-l} + (1-\theta_2) D_t$.
    \ELSE
    \STATE $\widehat{D}_t=\widehat{D}_{t-1}$.
    \ENDIF
    \STATE Compute the update step $\Delta x_{t}= - \widehat{D}_t^{-1} \hat{g}_t$.
    \STATE Compute the parameter-wise trust-region radius $$\eta=\text{SHD}(x_t,\epsilon).$$
    \STATE Apply parameter-wise clipping
    \begin{equation*}
      \widehat{\Delta} x_t = \Delta x_t.\text{clip}(-\eta,+\eta).  
    \end{equation*}
    \STATE Update the parameters $x_{t+1} = x_t + \widehat{\Delta} x_t$.
\ENDFOR
\end{algorithmic}
\end{algorithm}

\subsection{Algorithm Design}

We initialize the 3D Gaussian primitives using a sparse point cloud obtained from structure-from-motion (SfM). Given the parameters $x_t$ at iteration $t$, we first subsample a mini-batch of images $\mathcal{S}_1$ to estimate the stochastic gradient
\begin{equation*}
    g_t = \frac1m \frac{M}{|\mathcal{S}_1|}\sum_{i\in \mathcal{S}_1} J_i(x_t)^T f_i(x_t),
\end{equation*}
where $f_i(x_t)$ denotes the residual vector and $J_i(x_t)$ is its Jacobian of the $i$-th image with respect to the Gaussian parameters $x_t$. To reduce gradient noise and improve stability, we maintain an exponential moving average (EMA) of the gradients with decay rate $\theta_1 \in (0,1)$,
\begin{equation*}
    \widehat{g}_t = \theta_1 \widehat{g}_{t-1} + (1-\theta_1) g_t.
\end{equation*}

To incorporate second-order information while maintaining $O(n)$ memory and computation costs, we follow the spirit of Sophia \citep{liu2023sophia} and estimate the diagonal of the Gauss--Newton matrix $J^T J$ using Hutchinson’s method. Specifically, every $l$ iterations, we subsample another mini-batch of images $\mathcal{S}_2$ and compute
\begin{align*}
    D_t 
    &= \text{Hutch}(x_t, \mathcal{S}_2, \nu) \\
    &= \frac{1}{m}\frac{M}{|\mathcal{S}_2|}\frac1\nu\sum_{i=1}^{\nu} 
    z^{(i)} \odot 
    \left(
        \sum_{j\in \mathcal{S}_2} 
        J_j(x_t)^T J_j(x_t) z^{(i)}
    \right),
\end{align*}
where $\{z^{(i)}\}_{i=1}^m$ are independent Rademacher random vectors and $\odot$ denotes element-wise multiplication. The scaling factors $M|\mathcal{S}_1|^{-1}, M|\mathcal{S}_2|^{-1}$ are required to maintain an unbiased estimate of the gradient and the Hessian diagonal after sampling.
Such image subsampling and diagonal approximation significantly reduce computational and memory overhead,
since the Hessian-vector product $J_j(x_t)^TJ_j(x_t)z^{(i)}$ corresponds to exactly one forward and backward pass on a training image.
To further stabilize curvature estimates, we maintain an EMA of $D_t$ with decay rate $\theta_2$ and reuse the previous estimate in intermediate iterations, i.e.,
\begin{equation*}
    \widehat{D}_t = \left\{\begin{aligned}
    \theta_2 \widehat{D}_{t-1} + (1-\theta_2) D_t & \;\text{ if }t\mod l =1, \\
    \widehat{D}_{t-1} \quad\qquad & \;\text{ otherwise.}
    \end{aligned} \right.
\end{equation*}
Given $\widehat{D}_t$ and $\widehat{g}_t$, we compute the update step
\begin{equation*}
    \Delta x_t = - \widehat{D}_t^{-1} \widehat{g}_t.
\end{equation*}

To improve robustness under strong nonlinearity and discontinuities in the 3DGS rasterization process, we apply a parameter-wise trust-region constraint. Specifically, we compute a trust-region radius for each parameter
\begin{equation*}
    \eta = \text{SHD}(x_t,\epsilon),
\end{equation*}
based on the squared Hellinger distance, which bounds the change of each Gaussian primitive induced by the update. Details of this construction are provided in Section~\ref{sec:Trust-region based on squared Hellinger distance}. 

The final update is obtained by parameter-wise clipping,
\begin{equation*}
    \widehat{\Delta} x_t = \Delta x_t.\mathrm{clip}(-\eta, +\eta),
\end{equation*}
followed by the parameter update
\begin{equation}
    x_{t+1} = x_t + \widehat{\Delta} x_t.
\end{equation}

The minibatch sizes are selected based on the tradeoff between algorithmic performance and computational cost. In our experiments, we set $|\mathcal{S}_1|=|\mathcal{S}_2|=\nu=1$ and $l=10$, which corresponds to an approximately $10\%$ overhead compared to ADAM.

\subsection{Trust-Region Based on Squared Hellinger Distance}\label{sec:Trust-region based on squared Hellinger distance}

\begin{figure}[!t]
    \centering
    \includegraphics[width=0.8\linewidth]{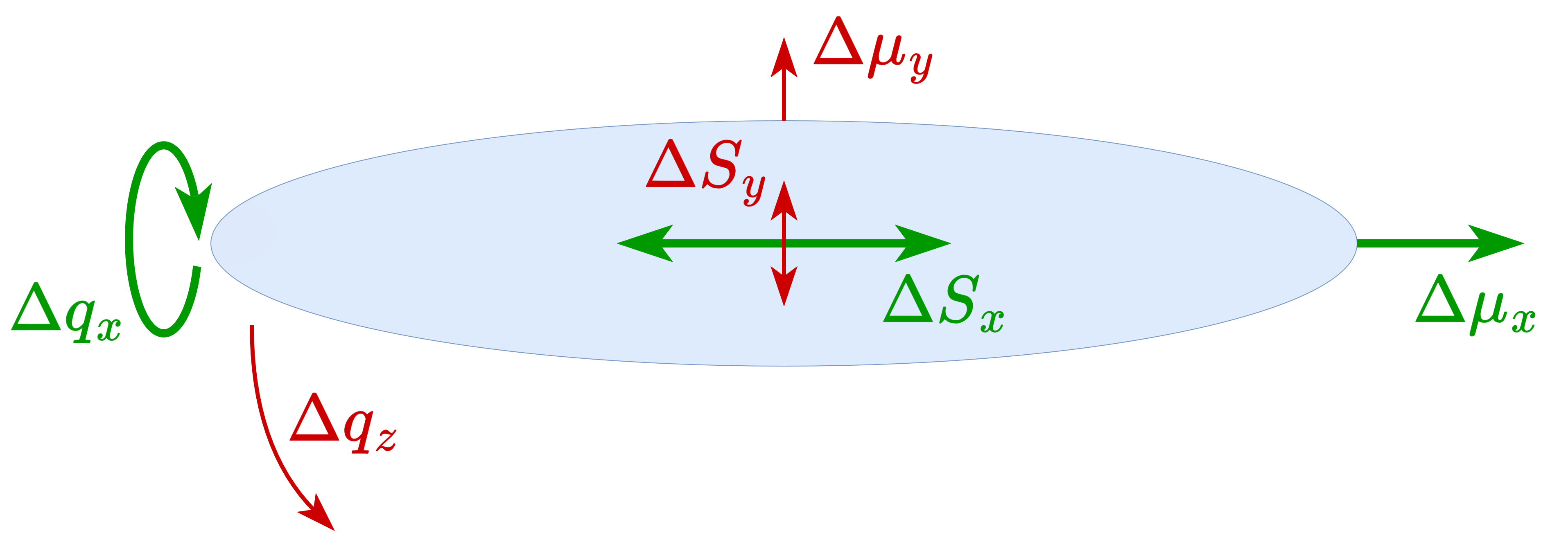}
    \caption{Example scene with a single elongated Gaussian splat. The $x$-axis and $y$-axis lie on the page; the $z$-axis comes out of the page. The green arrows mark the directions in which the Gaussian has more freedom; while the red arrows denote otherwise.}
    \label{fig:perturbation_diagram}
\end{figure}

As in Sophia~\citep{liu2023sophia}, individual update steps can occasionally become excessively large due to the high variance of Hutchinson’s estimator. 
This issue is further exacerbated in 3DGS by the highly nonlinear and piecewise-continuous nature of the rasterization process, particularly in later optimization stages where dense clusters of Gaussian primitives frequently emerge.

In ADAM~\citep{kingma2014adam}, this instability is typically addressed through careful learning-rate tuning and scheduled decay of the position updates. 
On the other hand, 3DGS-LM~\citep{hollein20253dgs} mitigates nonlinearity by introducing an adaptive regularization term that effectively interpolates between Gauss--Newton and gradient descent updates. However, both approaches operate at a global or group level and do not exploit the strong interpretability of individual Gaussian parameters.

We instead propose a \emph{parameter-wise trust-region} strategy that leverages the explicit geometric and photometric meaning of each parameter in 3DGS. Since interactions between Gaussian splats primarily occur through their projected opacity $\bar{\alpha}$ on the image plane, a natural trust-region design should directly limit changes to this quantity across viewpoints.

To build intuition, consider a scene containing a single anisotropic Gaussian (see \autoref{fig:perturbation_diagram}). Intuitively, the Gaussian should translate or expand more conservatively along its short axis to avoid abrupt changes in rendered pixels. 
Additionally, rotations about the long axis should be less restrictive than those about the short axis. Finally, more transparent Gaussians should be allowed to evolve more rapidly than highly opaque ones. All of these constraints can be formalized by constructing parameter-wise trust regions that bound the discrepancy between the Gaussian splat before and after an update.

\paragraph{Squared Hellinger distance.}
We quantify the magnitude of an update to a Gaussian $G$ using the squared Hellinger distance
\begin{align}
    H^2(G, G') = \int \left (\sqrt{G(z)} - \sqrt{G'(z)}\right)^2dz.
\end{align}
For unnormalized Gaussian primitives 
$G=Z \cdot \mathcal{N}(z;\mu,\Sigma)$ and $G'=Z' \cdot \mathcal{N}(z;\mu',\Sigma')$ with probability mass $Z, Z'$, $H^2(G, G')$ emits a closed-form solution
\begin{align*}
    H^2(G,G')  & = \frac12 (Z+Z') \\
    & \quad - (Z\cdot Z')^{\frac12} \cdot \Sigma_1 \cdot \exp(\Delta\mu^T\Sigma_2^{-1} \Delta\mu),
\end{align*}
where $\Delta\mu = \mu - \mu'$, $\Sigma_1=\frac{\det(\Sigma)^{\frac14} \det(\Sigma')^{\frac14}}{\det\left(\frac{\Sigma + \Sigma'}{2}\right)^{\frac12}}$, and
$\Sigma_2=\frac{\Sigma + \Sigma'}{2}$.

In the 3DGS setting, the probability mass satisfies
$Z= \alpha \mathcal{C} \det(\Sigma)^{1/2}$ and
$Z' = \alpha' \mathcal{C}' \det(\Sigma')^{1/2}$.
While alternative divergence measures (e.g., KL divergence) could also be employed, we find that the squared Hellinger distance yields trust-region bounds that are both simple to compute and intuitively interpretable, being the integral of the coordinate-wise difference between two Gaussian splats.

\paragraph{Scale normalization.}
Because the apparent mass of a rendered Gaussian can be arbitrarily scaled by moving along the viewing direction, we normalize the squared Hellinger distance by the determinant of the scale matrix. Specifically, we rescale $H^2(G, G')$ by
$\det(\Sigma)^{-1/2} = \det(S)^{-1}$,
recalling that $\Sigma = R^T S^T S R$ with orthogonal rotation matrix $R$.
This normalization can be interpreted as comparing Gaussians at a fixed effective distance to the camera. For all parameters except color, we treat the opacity $\alpha$ as the total mass of the Gaussian, as only the opacity component affects future Gaussians to be rendered.

\begin{figure*}[!t]
    \centering
    \includegraphics[width=0.98\linewidth]{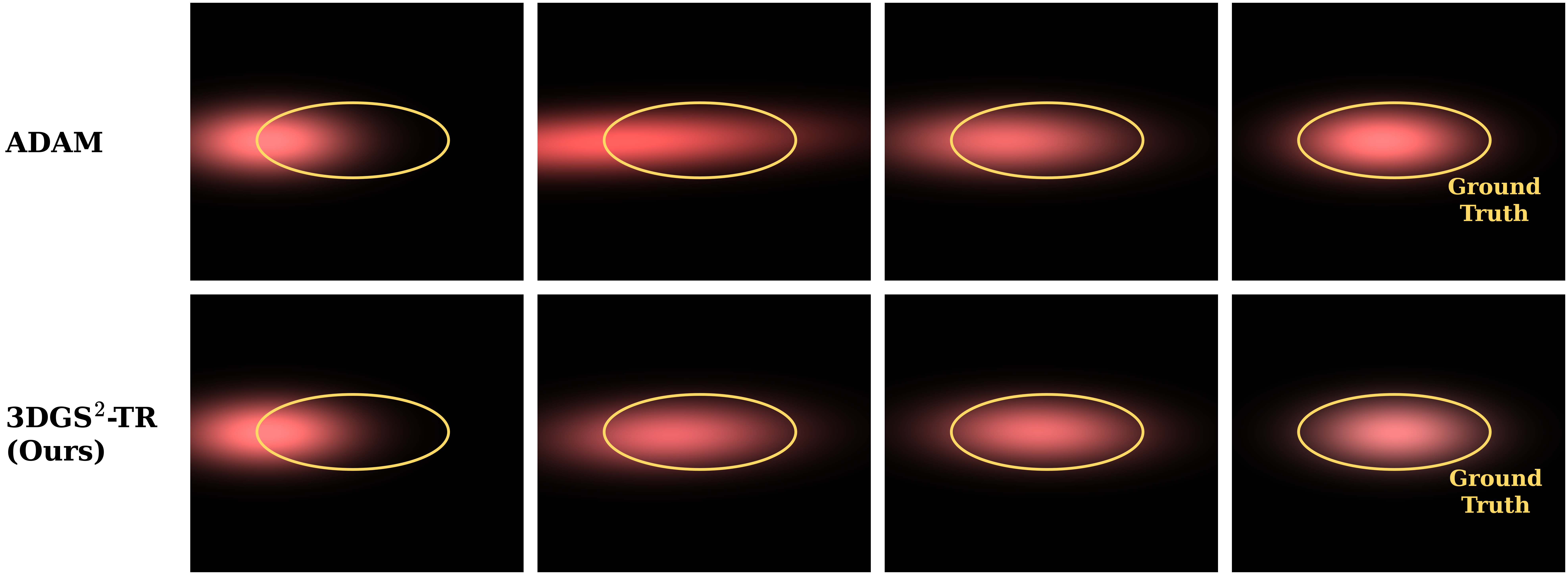}
    \caption{A single Gaussian fitting example where the Gaussian splat is initialized with a small perturbation.
    Images from left to right show the progression of optimization to fit the perturbed splat to the ground truth splat (denoted with the orange circle).
    Both ADAM and \project eventually recover the Gaussian parameters close to the ground truth, 
    but \project deforms the Gaussian much less due to the trust-region bound, which increases the stability of training.}
    \label{fig:gaussian-fitting-example}
\end{figure*}

\paragraph{Parameter-wise trust-region radius.}
We now derive a parameter-wise trust-region radius
$\eta = \mathrm{SHD}(x_t, \epsilon)$
such that, when a single parameter is perturbed while all others remain fixed, the normalized squared Hellinger distance between the Gaussian before the update $G(x_t)$ and after the update $G(x_t + \eta)$ is bounded by $\epsilon$:
\begin{equation*}
\eta = \mathrm{SHD}(x_t, \epsilon)
\coloneqq
\arg\max_{|\eta|}
\left\{
|\eta| :
\frac{H^2(G, G')}{\det(S)} < \epsilon
\right\}.
\end{equation*}
Below, we summarize the resulting trust-region bounds for each parameter type; detailed derivations are deferred to \autoref{sec:derivations}.

\paragraph{Bound on the mean $\mu=(\mu_x,\mu_y,\mu_z)$.}
For perturbations $\Delta\mu=(\Delta\mu_x,\Delta\mu_y,\Delta\mu_z)$, with $c\in\{x,y,z\}$, we require
\begin{align*}
   |\Delta\mu_c| &< \sqrt{-8\,\Sigma_{cc}\,\ln\!\left(1 - \frac{\epsilon}{\alpha}\right)}.
\end{align*}

\paragraph{Bound on the scale matrix $S$.}
Let
\begin{equation*}
  S=\begin{bmatrix} S_x && \\ & S_y & \\ && S_z \end{bmatrix},
\qquad
S'=\begin{bmatrix} S_x' && \\ & S_y' & \\ && S_z' \end{bmatrix},  
\end{equation*}
and define $\Delta S=S'-S$ with blockwise differences $\Delta S_x,\Delta S_y,\Delta S_z$, then with $c\in\{x,y,z\}$, we require
\begin{align*}
   |\Delta S_c| < \sqrt{\frac{2 S_c^2\,\epsilon}{\alpha}}.
\end{align*}

\paragraph{Bound on the opacity $\alpha$.}
Let $\alpha'=\alpha+\Delta\alpha$. To control the mass discrepancy, we require
\begin{equation*}
     |\Delta\alpha| < \sqrt{4\alpha\,\epsilon}.
\end{equation*}

\paragraph{Bound on the color $\mathcal{C}$.}
Let $\mathcal{C}=\mathcal{C}_r+\mathcal{C}_g+\mathcal{C}_b$,  $\mathcal{C}'=\mathcal{C}'_r+\mathcal{C}'_g+\mathcal{C}'_b$ and $\Delta\mathcal{C}_c=\mathcal{C}_c'-\mathcal{C}_c$ for $c\in\{r,g,b\}$. We impose
\begin{align*}
   |\Delta\mathcal{C}_c| < \sqrt{\frac{4\mathcal{C}_c\,\epsilon}{\alpha}}.
\end{align*}

\paragraph{Bound on the rotation $R$.}
The rotation matrix $R$ is parameterized by a quaternion $(q_x,q_y,q_z,q_w)$ with squared norm
$q^2 = q_x^2 + q_y^2 + q_z^2 + q_w^2$. Let $\beta_c > 0$ denote a geometry-dependent constant that upper bounds the sensitivity of the induced Gaussian displacement with respect to perturbations in the $c$-th quaternion component (see \autoref{sec:derivations}). 
To ensure the rotation update remains within the trust region, for $c \in \{x,y,z,w\}$ it suffices to require
\begin{align*}
   |\Delta q_c|
   < \sqrt{-\frac{8}{\beta_c}
   \ln\!\left(1 - \frac{\epsilon}{\alpha}\right)}.
\end{align*}


We illustrate the effects of the proposed Hellinger-distance-based trust region using a single Gaussian fitting example in \autoref{fig:gaussian-fitting-example}.

\section{Evaluations}

We now present the evaluation results of our method. 
The forward-mode automatic differentiation of the 3DGS rasterization kernel is implemented in CUDA, 
based on an existing third-party implementation.  
The parameter-wise trust-region radius is implemented as a custom CUDA kernel.
The training pipeline and the Sophia optimizer is implemented in PyTorch following the original 3DGS implementation.

Our experiments are run on A100-SXM4 GPUs with 6921 CUDA cores and 80 GB VRAM.
We evaluate our method on the same datasets as 3DGS, namely, all scenes from MiP-NeRF360, two scenes from Tanks \& Temples, and two scenes from Deep Blending \cite{barron2022mipnerf360, Knapitsch2017, DeepBlending2018}.
For each dataset, we initialize the 3D Gaussian splats with the standard SfM point cloud.
As our diagonal estimator currently does not handle inserting new diagonal entries, all of our experiments are performed \textit{without densification.} 

 In all experiments, we set the Hessian diagonal update interval to $l=10$ to balance computational cost and curvature accuracy,
following the default configuration of Sophia~\citep{liu2023sophia}, 
and choose the exponential moving average (EMA) parameters $\theta_1=0.9$ and $\theta_2=0.999$. 
The trust-region radius $\epsilon$ follows a exponential decay schedule from $10^{-6}$ to $10^{-8}$ over the course of training.

For our baselines, we compare our method against ADAM, the SOTA first-order method, 
and 3DGS-LM, the only second-order optimizer for 3DGS with open-source implementation. 

\begin{table*}[t]
\resizebox{\textwidth}{!}{
\begin{threeparttable}
\centering
\caption{The main quantitative result. Best results are \textbf{boldfaced} and second best are \underline{underlined}.}
\label{tab:summary_table}
\begin{tabular}{lccccccccccc}
\toprule
    \midrule
    \multirow{2}{*}{Method}
      & \multicolumn{3}{c}{SSIM $\downarrow$}
      & \multicolumn{3}{c}{PSNR $\uparrow$}
      & \multicolumn{3}{c}{LPIPS $\downarrow$} 
      & \multirow{2}{*}{\makecell{Time\\(s)}}
      & \multirow{2}{*}{\makecell{Peak GPU\\Mem (GB)}} \\
    \cmidrule(lr){2-4}
    \cmidrule(lr){5-7}
    \cmidrule(lr){8-10}
      & @7k & @15k & @30k
      & @7k & @15k & @30k
      & @7k & @15k & @30k & & \\
    \midrule

\multicolumn{12}{c}{mipnerf} \\
\midrule
ADAM & 0.667 & 0.679 & 0.684 & 24.45 & 24.85 & 25.08 & 0.406 & 0.400 & 0.420 & \textbf{212} & \textbf{5.60} \\
3DGS-LM \footnotemark[1] & 0.493 & 0.520 & 0.549 & 20.02 & 21.02 & 21.88 & 0.598 & 0.577 & 0.548 & 2602 & 47.75\\
ADAM-TR (Ours) & \underline{0.674} & \underline{0.688} & \underline{0.693} & \underline{24.54} & \underline{25.02} & \underline{25.21} & \underline{0.394} & \underline{0.388} & \underline{0.411} & \underline{235} & \underline{5.64}\\
3DGS$^2$-TR (Ours) \footnotemark[2] & \textbf{0.682} & \textbf{0.692} & \textbf{0.696} & \textbf{24.80} & \textbf{25.19} & \textbf{25.39} & \textbf{0.390} & \textbf{0.385} & \textbf{0.402} & 652 & 6.51 \\
\midrule

\multicolumn{12}{c}{deepblending} \\
\midrule
ADAM & 0.845 & 0.854 & 0.857 & \underline{26.65} & 27.18 & 27.36 & 0.372 & 0.365 & 0.387 & \textbf{180} & \textbf{5.42} \\
3DGS-LM \footnotemark[1] & 0.810 & 0.833 & 0.845 & 23.72 & 24.89 & 25.62 & 0.437 & 0.405 & 0.385 & 2235 & 33.33\\
ADAM-TR (Ours) & \underline{0.855} & \underline{0.866} & \textbf{0.869} & 26.56 & \underline{27.34} & \underline{27.58} & \underline{0.357} & \underline{0.351} & \underline{0.375} & \underline{195} & \underline{5.51}\\
3DGS$^2$-TR (Ours) \footnotemark[2] & \textbf{0.859} & \textbf{0.867} & \underline{0.869} & \textbf{26.91} & \textbf{27.54} & \textbf{27.74} & \textbf{0.354} & \textbf{0.349} & \textbf{0.367} & 600 & 6.26 \\
\midrule

\multicolumn{12}{c}{tandt} \\
\midrule
ADAM & 0.723 & 0.755 & 0.766 & 20.69 & 21.41 & 21.66 & 0.309 & 0.298 & 0.341 & \textbf{201} & \textbf{3.17} \\
3DGS-LM \footnotemark[1] & 0.661 & 0.717 & 0.749 & 20.00 & 21.09 & 21.81 & 0.429 & 0.366 & 0.331 & 2105 & 25.53\\
ADAM-TR (Ours) & \underline{0.774} & \underline{0.791} & \underline{0.798} & \underline{21.62} & \underline{22.16} & \underline{22.42} & \underline{0.273} & \underline{0.264} & \underline{0.295} & \underline{227} & \underline{3.27}\\
3DGS$^2$-TR (Ours) \footnotemark[2] & \textbf{0.787} & \textbf{0.800} & \textbf{0.805} & \textbf{22.12} & \textbf{22.63} & \textbf{22.85} & \textbf{0.262} & \textbf{0.255} & \textbf{0.279} & 484 & 4.12 \\
\midrule
\bottomrule
\end{tabular}
\begin{tablenotes}
\item[1] 3DGS-LM results are reported @35, @75, and @150 iterations.
\item[2] System implementation is not fully optimized. See Section~\ref{sec:limitations} for details.
\end{tablenotes}
\end{threeparttable}
}
\end{table*}

\begin{figure*}[!t]
    \centering
    \includegraphics[width=0.98\linewidth]{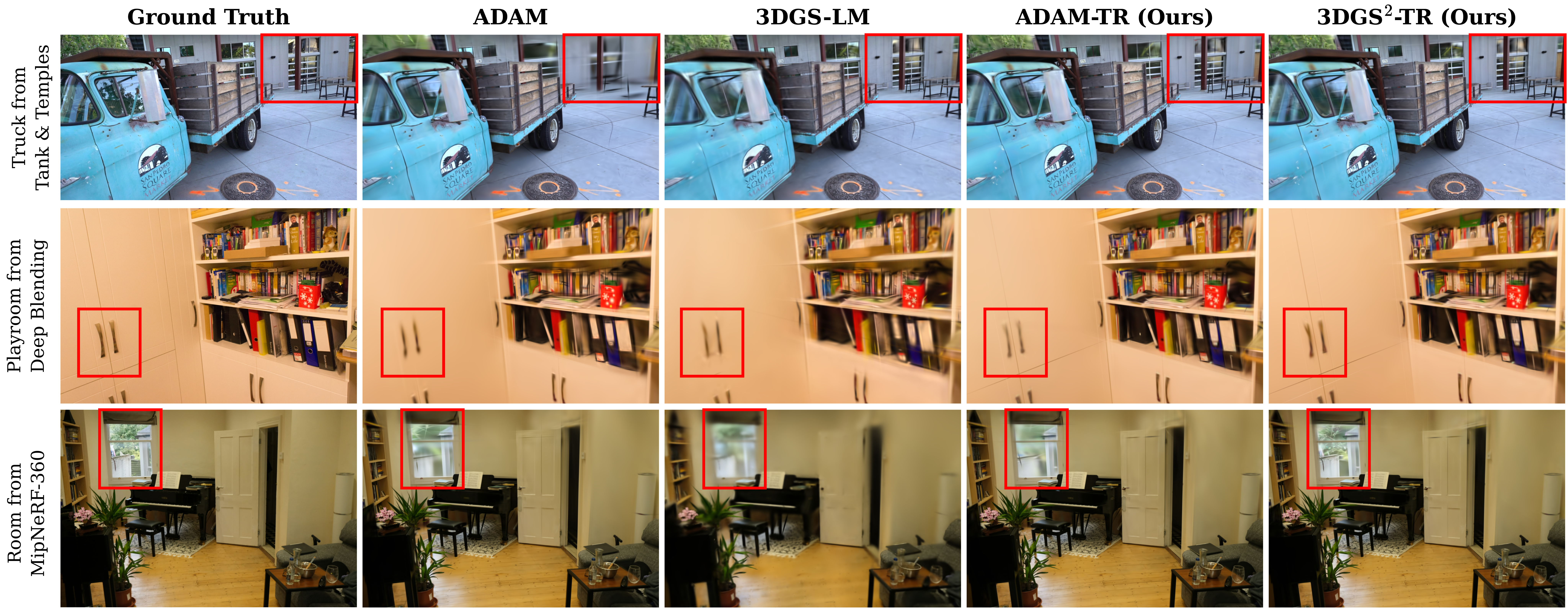}
    \caption{Qualitative comparison of different methods of the truck, playroom, and room scenes. 
    The red boxes highlight regions where \project significantly outperforms other methods.}
\label{fig:qualitative_comparison}
\end{figure*}

We also include an ablation study which applies our Hellinger-distance-based trust region to the ADAM update step (ADAM-TR).

Since 3DGS-LM has a high computation overhead, we only run it for 150 iterations in total, 
and report the results at 35, 75, and 150 iterations.
We run all other methods for 30k iterations and report results at 7k, 15k, and 30k iterations.
We use the same train/test split and report the same metrics (SSIM, PSNR, LPIPS) on the test images as proposed by 3DGS.

The main quantitative results are presented in \autoref{tab:summary_table}. 
(Results for individual scenes can be found in \autoref{sec:results_per_scene}.)
Qualitative comparisons for some scenes are shown in \autoref{fig:qualitative_comparison}. 
 \project significantly outperforms other methods, reaching comparable or better reconstruction quality using 50\% fewer iterations than first-order methods (ADAM and ADAM-TR). In the case of Tanks \& Temples, \project is able to exceed the best ADAM PSNR by 0.56dB at 7k iterations and 1.19dB at 30k iterations.
Moreover, \project requires less than 1GB of additional GPU memory for training in all scenes, which is 17\% more than ADAM and 85\% less than 3DGS-LM on average. 

In contrast, 3DGS-LM does not produce comparable results with the same initialization, given significantly more time and resources.
We also note that by applying our trust-region clipping to the ADAM updates (ADAM-TR), we are already able to achieve better quality than the vanilla ADAM optimizer.

\section{Discussions and Limitations}

\label{sec:limitations}

Our method achieves much faster convergence per step compared to ADAM. 
However, due to a naive implementation of the Sophia optimizer in PyTorch, our second-order method has a higher run time per iteration, which is not a fundamental limitation of the algorithm itself. 
We plan to optimize the PyTorch implementation to reach comparable performance to ADAM for tensor manipulation in future work.

Compared to other proposed second-order optimizers, our method is fairly non-intrusive to the vanilla training pipeline, 
which makes it amenable to orthogonal improvements to 3DGS training, such as Markov Chain Monte Carlo (MCMC) densification \cite{kheradmand20243d}, co-regularization \cite{zhang2024cor}, drop-out Gaussians \cite{park2025dropgaussian}, etc.
However, one current limitation with our work is that we do not support adding or moving around Gaussian splats. 
Naively inserting rows and columns to the Hessian matrix results in a biased estimation of the diagonal entries, 
leading to degraded performance, which we aim to resolve in future work.

\balance
\section{Conclusion}

We present \project, a second-order optimizer for training 3D Gaussian Splatting scenes. 
We approximate the curvature information of the loss function using a stochastic diagonal estimator, which eliminates the storage and computation overhead of classical second-order methods, thus maintains the same $O(n)$ complexity as ADAM in both computational cost and
memory. 
Furthermore, we introduce a parameter-wise trust-region technique  based on the squared Hellinger distance to bound the update step size. Instead of tuning learning rates for each parameter group, our method requires only one hyperparameter. We show that \project achieves better reconstruction quality after 50\% of training iterations, while incurring only 10\% of computation overhead and less than 1GB of memory overhead compared to ADAM. 
Our method is a drop-in replacement for the ADAM optimizer to accelerate 3DGS training in all settings.

\newpage
\section*{Impact Statement}
This paper presents work whose goal is to advance the field of Machine
Learning. There are many potential societal consequences of our work, none
which we feel must be specifically highlighted here.

\bibliography{main}

@String(CVPR= {IEEE Conf. Comput. Vis. Pattern Recog.})

@String(TOG= {ACM Trans. Graph.})

@String(AAAI = {AAAI})

@String(CVPR  = {CVPR})

@String(TOG   = {ACM TOG})

@article{pehlivan2025second,
  title={Second-order Optimization of Gaussian Splats with Importance Sampling},
  author={Pehlivan, Hamza and Camiletto, Andrea Boscolo and Foo, Lin Geng and Habermann, Marc and Theobalt, Christian},
  journal={arXiv preprint arXiv:2504.12905},
  year={2025}
}

@article{liu2023sophia,
  title={Sophia: A scalable stochastic second-order optimizer for language model pre-training},
  author={Liu, Hong and Li, Zhiyuan and Hall, David and Liang, Percy and Ma, Tengyu},
  journal={arXiv preprint arXiv:2305.14342},
  year={2023}
}

@article{dauphin2014identifying,
  title={Identifying and attacking the saddle point problem in high-dimensional non-convex optimization},
  author={Dauphin, Yann N and Pascanu, Razvan and Gulcehre, Caglar and Cho, Kyunghyun and Ganguli, Surya and Bengio, Yoshua},
  journal={Advances in neural information processing systems},
  volume={27},
  year={2014}
}

@inproceedings{yao2021adahessian,
  title={Adahessian: An adaptive second order optimizer for machine learning},
  author={Yao, Zhewei and Gholami, Amir and Shen, Sheng and Mustafa, Mustafa and Keutzer, Kurt and Mahoney, Michael},
  booktitle={proceedings of the AAAI conference on artificial intelligence},
  volume={35},
  number={12},
  pages={10665--10673},
  year={2021}
}

@InProceedings{schaul13no,
  title = 	 {No more pesky learning rates},
  author = 	 {Schaul, Tom and Zhang, Sixin and LeCun, Yann},
  booktitle = 	 {Proceedings of the 30th International Conference on Machine Learning},
  pages = 	 {343--351},
  year = 	 {2013}
}

@inproceedings{sutton1986two,
  title={Two problems with backpropagation and other steepest-descent learning procedures for networks},
  author={Sutton, Richard S},
  booktitle={Proceedings of the annual meeting of the cognitive science society},
  volume={8},
  year={1986}
}

@article{mildenhall2021nerf,
  title={Nerf: Representing scenes as neural radiance fields for view synthesis},
  author={Mildenhall, Ben and Srinivasan, Pratul P and Tancik, Matthew and Barron, Jonathan T and Ramamoorthi, Ravi and Ng, Ren},
  journal={Communications of the ACM},
  volume={65},
  number={1},
  pages={99--106},
  year={2021},
  publisher={ACM New York, NY, USA}
}

@article{kerbl20233d,
  title={3D Gaussian splatting for real-time radiance field rendering.},
  author={Kerbl, Bernhard and Kopanas, Georgios and Leimk{\"u}hler, Thomas and Drettakis, George},
  journal={ACM Trans. Graph.},
  volume={42},
  number={4},
  pages={139--1},
  year={2023}
}

@article{kingma2014adam,
  title={Adam: A method for stochastic optimization},
  author={Kingma, Diederik P},
  journal={arXiv preprint arXiv:1412.6980},
  year={2014}
}

@article{bottou2018optimization,
  title={Optimization methods for large-scale machine learning},
  author={Bottou, L{\'e}on and Curtis, Frank E and Nocedal, Jorge},
  journal={SIAM review},
  volume={60},
  number={2},
  pages={223--311},
  year={2018},
  publisher={SIAM}
}

@article{durvasula2023distwar,
  title={Distwar: Fast differentiable rendering on raster-based rendering pipelines},
  author={Durvasula, Sankeerth and Zhao, Adrian and Chen, Fan and Liang, Ruofan and Sanjaya, Pawan Kumar and Vijaykumar, Nandita},
  journal={arXiv preprint arXiv:2401.05345},
  year={2023}
}

@inproceedings{feng2025flashgs,
  title={Flashgs: Efficient 3d gaussian splatting for large-scale and high-resolution rendering},
  author={Feng, Guofeng and Chen, Siyan and Fu, Rong and Liao, Zimu and Wang, Yi and Liu, Tao and Hu, Boni and Xu, Linning and Pei, Zhilin and Li, Hengjie and others},
  booktitle={Proceedings of the Computer Vision and Pattern Recognition Conference},
  pages={26652--26662},
  year={2025}
}

@inproceedings{mallick2024taming,
  title={Taming 3dgs: High-quality radiance fields with limited resources},
  author={Mallick, Saswat Subhajyoti and Goel, Rahul and Kerbl, Bernhard and Steinberger, Markus and Carrasco, Francisco Vicente and De La Torre, Fernando},
  booktitle={SIGGRAPH Asia 2024 Conference Papers},
  pages={1--11},
  year={2024}
}

@article{ye2025gsplat,
  title={gsplat: An open-source library for Gaussian splatting},
  author={Ye, Vickie and Li, Ruilong and Kerr, Justin and Turkulainen, Matias and Yi, Brent and Pan, Zhuoyang and Seiskari, Otto and Ye, Jianbo and Hu, Jeffrey and Tancik, Matthew and others},
  journal={Journal of Machine Learning Research},
  volume={26},
  number={34},
  pages={1--17},
  year={2025}
}

@article{kheradmand20243d,
  title={3d gaussian splatting as markov chain monte carlo},
  author={Kheradmand, Shakiba and Rebain, Daniel and Sharma, Gopal and Sun, Weiwei and Tseng, Yang-Che and Isack, Hossam and Kar, Abhishek and Tagliasacchi, Andrea and Yi, Kwang Moo},
  journal={Advances in Neural Information Processing Systems},
  volume={37},
  pages={80965--80986},
  year={2024}
}

@inproceedings{fang2024mini,
  title={Mini-splatting: Representing scenes with a constrained number of gaussians},
  author={Fang, Guangchi and Wang, Bing},
  booktitle={European Conference on Computer Vision},
  pages={165--181},
  year={2024},
  organization={Springer}
}

@inproceedings{rota2024revising,
  title={Revising densification in gaussian splatting},
  author={Rota Bul{\`o}, Samuel and Porzi, Lorenzo and Kontschieder, Peter},
  booktitle={European Conference on Computer Vision},
  pages={347--362},
  year={2024},
  organization={Springer}
}

@article{zhang2024lp,
  title={Lp-3dgs: Learning to prune 3d gaussian splatting},
  author={Zhang, Zhaoliang and Song, Tianchen and Lee, Yongjae and Yang, Li and Peng, Cheng and Chellappa, Rama and Fan, Deliang},
  journal={Advances in Neural Information Processing Systems},
  volume={37},
  pages={122434--122457},
  year={2024}
}

@inproceedings{hanson2025pup,
  title={Pup 3d-gs: Principled uncertainty pruning for 3d gaussian splatting},
  author={Hanson, Alex and Tu, Allen and Singla, Vasu and Jayawardhana, Mayuka and Zwicker, Matthias and Goldstein, Tom},
  booktitle={Proceedings of the Computer Vision and Pattern Recognition Conference},
  pages={5949--5958},
  year={2025}
}

@article{ali2024trimming,
  title={Trimming the fat: Efficient compression of 3d gaussian splats through pruning},
  author={Ali, Muhammad Salman and Qamar, Maryam and Bae, Sung-Ho and Tartaglione, Enzo},
  journal={arXiv preprint arXiv:2406.18214},
  year={2024}
}

@inproceedings{hollein20253dgs,
  title={3dgs-lm: Faster gaussian-splatting optimization with levenberg-marquardt},
  author={H{\"o}llein, Lukas and Bo{\v{z}}i{\v{c}}, Alja{\v{z}} and Zollh{\"o}fer, Michael and Nie{\ss}ner, Matthias},
  booktitle={Proceedings of the IEEE/CVF International Conference on Computer Vision},
  pages={26740--26750},
  year={2025}
}

@inproceedings{wang2024adr,
  title={Adr-gaussian: Accelerating gaussian splatting with adaptive radius},
  author={Wang, Xinzhe and Yi, Ran and Ma, Lizhuang},
  booktitle={SIGGRAPH Asia 2024 Conference Papers},
  pages={1--10},
  year={2024}
}

@inproceedings{hanson2025speedy,
  title={Speedy-splat: Fast 3d gaussian splatting with sparse pixels and sparse primitives},
  author={Hanson, Alex and Tu, Allen and Lin, Geng and Singla, Vasu and Zwicker, Matthias and Goldstein, Tom},
  booktitle={Proceedings of the Computer Vision and Pattern Recognition Conference},
  pages={21537--21546},
  year={2025}
}

@inproceedings{lee2024compact,
  title={Compact 3d gaussian representation for radiance field},
  author={Lee, Joo Chan and Rho, Daniel and Sun, Xiangyu and Ko, Jong Hwan and Park, Eunbyung},
  booktitle={Proceedings of the IEEE/CVF Conference on Computer Vision and Pattern Recognition},
  pages={21719--21728},
  year={2024}
}

@article{fan2024lightgaussian,
  title={Lightgaussian: Unbounded 3d gaussian compression with 15x reduction and 200+ fps},
  author={Fan, Zhiwen and Wang, Kevin and Wen, Kairun and Zhu, Zehao and Xu, Dejia and Wang, Zhangyang and others},
  journal={Advances in neural information processing systems},
  volume={37},
  pages={140138--140158},
  year={2024}
}

@inproceedings{girish2024eagles,
  title={Eagles: Efficient accelerated 3d gaussians with lightweight encodings},
  author={Girish, Sharath and Gupta, Kamal and Shrivastava, Abhinav},
  booktitle={European Conference on Computer Vision},
  pages={54--71},
  year={2024},
  organization={Springer}
}

@inproceedings{song2024city,
  title={City-on-web: Real-time neural rendering of large-scale scenes on the web},
  author={Song, Kaiwen and Zeng, Xiaoyi and Ren, Chenqu and Zhang, Juyong},
  booktitle={European Conference on Computer Vision},
  pages={385--402},
  year={2024},
  organization={Springer}
}

@article{kerbl2024hierarchical,
  title={A hierarchical 3d gaussian representation for real-time rendering of very large datasets},
  author={Kerbl, Bernhard and Meuleman, Andreas and Kopanas, Georgios and Wimmer, Michael and Lanvin, Alexandre and Drettakis, George},
  journal={ACM Transactions on Graphics (TOG)},
  volume={43},
  number={4},
  pages={1--15},
  year={2024},
  publisher={ACM New York, NY, USA}
}

@inproceedings{yu2024mip,
  title={Mip-splatting: Alias-free 3d gaussian splatting},
  author={Yu, Zehao and Chen, Anpei and Huang, Binbin and Sattler, Torsten and Geiger, Andreas},
  booktitle={Proceedings of the IEEE/CVF conference on computer vision and pattern recognition},
  pages={19447--19456},
  year={2024}
}

@inproceedings{lu2024scaffold,
  title={Scaffold-gs: Structured 3d gaussians for view-adaptive rendering},
  author={Lu, Tao and Yu, Mulin and Xu, Linning and Xiangli, Yuanbo and Wang, Limin and Lin, Dahua and Dai, Bo},
  booktitle={Proceedings of the IEEE/CVF Conference on Computer Vision and Pattern Recognition},
  pages={20654--20664},
  year={2024}
}

@inproceedings{lan20253dgs2,
  title={3dgs2: Near second-order converging 3d gaussian splatting},
  author={Lan, Lei and Shao, Tianjia and Lu, Zixuan and Zhang, Yu and Jiang, Chenfanfu and Yang, Yin},
  booktitle={Proceedings of the Special Interest Group on Computer Graphics and Interactive Techniques Conference Conference Papers},
  pages={1--10},
  year={2025}
}

@inproceedings{zhao2024scaling,
  title={On scaling up 3d gaussian splatting training},
  author={Zhao, Hexu and Weng, Haoyang and Lu, Daohan and Li, Ang and Li, Jinyang and Panda, Aurojit and Xie, Saining},
  booktitle={European Conference on Computer Vision},
  pages={14--36},
  year={2024},
  organization={Springer}
}

@article{papantonakis2024reducing,
  title={Reducing the memory footprint of 3d gaussian splatting},
  author={Papantonakis, Panagiotis and Kopanas, Georgios and Kerbl, Bernhard and Lanvin, Alexandre and Drettakis, George},
  journal={Proceedings of the ACM on Computer Graphics and Interactive Techniques},
  volume={7},
  number={1},
  pages={1--17},
  year={2024},
  publisher={ACM New York, NY, USA}
}

@article{barron2022mipnerf360,
    title={Mip-NeRF 360: Unbounded Anti-Aliased Neural Radiance Fields},
    author={Jonathan T. Barron and Ben Mildenhall and 
            Dor Verbin and Pratul P. Srinivasan and Peter Hedman},
    journal={CVPR},
    year={2022}
}

@article{Knapitsch2017,
    author    = {Arno Knapitsch and Jaesik Park and Qian-Yi Zhou and Vladlen Koltun},
    title     = {Tanks and Temples: Benchmarking Large-Scale Scene Reconstruction},
    journal   = {ACM Transactions on Graphics},
    volume    = {36},
    number    = {4},
    year      = {2017},
}

@article{DeepBlending2018,
  author = {Hedman, Peter and Philip, Julien and Price, True and Frahm, Jan-Michael and Drettakis, George and Brostow, Gabriel},
  title = {Deep Blending for Free-viewpoint Image-based Rendering},
  booktitle = {ACM Transactions on Graphics (Proc. SIGGRAPH Asia)},
  publisher = {ACM},
  volume    = {37},
  number    = {6},
  pages     = {257:1--257:15},
  year      = {2018}
}

@inproceedings{zhang2024cor,
  title={Cor-gs: sparse-view 3d gaussian splatting via co-regularization},
  author={Zhang, Jiawei and Li, Jiahe and Yu, Xiaohan and Huang, Lei and Gu, Lin and Zheng, Jin and Bai, Xiao},
  booktitle={European Conference on Computer Vision},
  pages={335--352},
  year={2024},
  organization={Springer}
}

@inproceedings{park2025dropgaussian,
  title={Dropgaussian: Structural regularization for sparse-view gaussian splatting},
  author={Park, Hyunwoo and Ryu, Gun and Kim, Wonjun},
  booktitle={Proceedings of the Computer Vision and Pattern Recognition Conference},
  pages={21600--21609},
  year={2025}
}

\nocite{langley00}

\bibliographystyle{icml2026}

\newpage
\appendix
\onecolumn
\section{Quantitative Results Per Scene}\label{sec:results_per_scene}

\begin{table*}[h]
\resizebox{\textwidth}{!}{
\begin{threeparttable}
\centering
\caption{Results on Deep Blending. Best results are \textbf{boldfaced} and second best are \underline{underlined}.}
\label{tab:deepblending}
\begin{tabular}{llccccccccccc}
\toprule
\midrule
\multirow{3}{*}{Method} & \multirow{3}{*}{Scene}
  & \multicolumn{11}{c}{Deep Blending} \\
\cmidrule(lr){3-13}
  & 
  & \multicolumn{3}{c}{SSIM $\downarrow$}
  & \multicolumn{3}{c}{PSNR $\uparrow$}
  & \multicolumn{3}{c}{LPIPS $\downarrow$} 
  & \multirow{2}{*}{\makecell{Time\\(s)}}
  & \multirow{2}{*}{\makecell{Peak GPU\\Mem (GB)}} \\
\cmidrule(lr){3-5}
\cmidrule(lr){6-8}
\cmidrule(lr){9-11}
  & 
  & @7k & @15k & @30k
  & @7k & @15k & @30k
  & @7k & @15k & @30k & & \\
\midrule

ADAM & drjohnson & 0.841 & 0.855 & 0.861 & \textbf{26.60} & \textbf{27.39} & \textbf{27.64} & 0.368 & 0.360 & 0.390 & \textbf{182} & \textbf{5.82} \\
3DGS-LM \footnotemark[1] & drjohnson & 0.811 & 0.837 & 0.849 & 24.78 & 25.87 & 26.37 & 0.432 & 0.393 & \underline{0.373} & 3344 & 47.65\\
ADAM-TR (Ours) & drjohnson & \underline{0.852} & \textbf{0.865} & \textbf{0.869} & 26.50 & \underline{27.30} & \underline{27.56} & \underline{0.355} & \underline{0.347} & 0.378 & \underline{200} & \underline{5.90}\\
3DGS$^2$-TR (Ours) \footnotemark[2] & drjohnson & \textbf{0.855} & \underline{0.865} & \underline{0.869} & \underline{26.54} & 27.25 & 27.48 & \textbf{0.353} & \textbf{0.346} & \textbf{0.371} & 606 & 6.68 \\
\midrule

ADAM & playroom & 0.849 & 0.852 & 0.854 & \underline{26.70} & 26.97 & 27.07 & 0.376 & 0.371 & 0.385 & \textbf{177} & \textbf{5.03} \\
3DGS-LM \footnotemark[1] & playroom & 0.809 & 0.829 & 0.840 & 22.66 & 23.91 & 24.87 & 0.443 & 0.417 & 0.398 & 1125 & 19.01\\
ADAM-TR (Ours) & playroom & \underline{0.859} & \underline{0.867} & \underline{0.869} & 26.63 & \underline{27.38} & \underline{27.61} & \underline{0.359} & \underline{0.354} & \underline{0.373} & \underline{189} & \underline{5.12}\\
3DGS$^2$-TR (Ours) \footnotemark[2] & playroom & \textbf{0.863} & \textbf{0.868} & \textbf{0.870} & \textbf{27.28} & \textbf{27.84} & \textbf{27.99} & \textbf{0.354} & \textbf{0.351} & \textbf{0.363} & 594 & 5.84 \\
\midrule
\bottomrule
    \end{tabular}
    \begin{tablenotes}
    \item[1] 3DGS-LM results are reported @35, @75, and @150 iterations.
    \item[2] System implementation is not fully optimized. See Section~\ref{sec:limitations} for details.
    \end{tablenotes}
    \end{threeparttable}
    }
    \end{table*}

\begin{table*}[h]
\resizebox{\textwidth}{!}{
\begin{threeparttable}
\centering
\caption{Results on Tanks \& Temples. Best results are \textbf{boldfaced} and second best are \underline{underlined}.}
\label{tab:tandt}
\begin{tabular}{llccccccccccc}
\toprule
\midrule
\multirow{3}{*}{Method} & \multirow{3}{*}{Scene}
  & \multicolumn{11}{c}{Tanks \& Temples} \\
\cmidrule(lr){3-13}
  & 
  & \multicolumn{3}{c}{SSIM $\downarrow$}
  & \multicolumn{3}{c}{PSNR $\uparrow$}
  & \multicolumn{3}{c}{LPIPS $\downarrow$} 
  & \multirow{2}{*}{\makecell{Time\\(s)}}
  & \multirow{2}{*}{\makecell{Peak GPU\\Mem (GB)}} \\
\cmidrule(lr){3-5}
\cmidrule(lr){6-8}
\cmidrule(lr){9-11}
  & 
  & @7k & @15k & @30k
  & @7k & @15k & @30k
  & @7k & @15k & @30k & & \\
\midrule

ADAM & train & 0.697 & 0.745 & 0.758 & 19.83 & \underline{20.91} & 21.21 & 0.318 & 0.302 & 0.362 & \textbf{209} & \textbf{3.45} \\
3DGS-LM \footnotemark[1] & train & 0.644 & 0.688 & 0.720 & 19.13 & 19.92 & 20.57 & 0.425 & 0.377 & 0.347 & 2162 & 28.36\\
ADAM-TR (Ours) & train & \underline{0.740} & \underline{0.763} & \underline{0.771} & \underline{20.21} & 20.87 & \underline{21.22} & \underline{0.297} & \underline{0.287} & \underline{0.321} & \underline{229} & \underline{3.54}\\
3DGS$^2$-TR (Ours) \footnotemark[2] & train & \textbf{0.757} & \textbf{0.775} & \textbf{0.781} & \textbf{20.68} & \textbf{21.32} & \textbf{21.62} & \textbf{0.283} & \textbf{0.276} & \textbf{0.302} & 503 & 4.43 \\
\midrule

ADAM & truck & 0.750 & 0.766 & 0.773 & 21.56 & 21.90 & 22.12 & 0.301 & 0.293 & 0.320 & \textbf{192} & \textbf{2.88} \\
3DGS-LM \footnotemark[1] & truck & 0.679 & 0.747 & 0.779 & 20.86 & 22.25 & 23.05 & 0.432 & 0.355 & 0.315 & 2047 & 22.69\\
ADAM-TR (Ours) & truck & \underline{0.807} & \underline{0.819} & \underline{0.824} & \underline{23.02} & \underline{23.44} & \underline{23.62} & \underline{0.250} & \underline{0.242} & \underline{0.268} & \underline{224} & \underline{3.00}\\
3DGS$^2$-TR (Ours) \footnotemark[2] & truck & \textbf{0.816} & \textbf{0.826} & \textbf{0.829} & \textbf{23.55} & \textbf{23.93} & \textbf{24.07} & \textbf{0.240} & \textbf{0.234} & \textbf{0.255} & 465 & 3.82 \\
\midrule
\bottomrule
    \end{tabular}
    \begin{tablenotes}
    \item[1] 3DGS-LM results are reported @35, @75, and @150 iterations.
    \item[2] System implementation is not fully optimized. See Section~\ref{sec:limitations} for details.
    \end{tablenotes}
    \end{threeparttable}
    }
    \end{table*}

\begin{table*}[h]
\resizebox{\textwidth}{!}{
\begin{threeparttable}
\centering
\caption{Results on MipNeRF-360. Best results are \textbf{boldfaced} and second best are \underline{underlined}.}
\label{tab:mipnerf}
\begin{tabular}{llccccccccccc}
\toprule
\midrule
\multirow{3}{*}{Method} & \multirow{3}{*}{Scene}
  & \multicolumn{11}{c}{MipNeRF-360} \\
\cmidrule(lr){3-13}
  & 
  & \multicolumn{3}{c}{SSIM $\downarrow$}
  & \multicolumn{3}{c}{PSNR $\uparrow$}
  & \multicolumn{3}{c}{LPIPS $\downarrow$} 
  & \multirow{2}{*}{\makecell{Time\\(s)}}
  & \multirow{2}{*}{\makecell{Peak GPU\\Mem (GB)}} \\
\cmidrule(lr){3-5}
\cmidrule(lr){6-8}
\cmidrule(lr){9-11}
  & 
  & @7k & @15k & @30k
  & @7k & @15k & @30k
  & @7k & @15k & @30k & & \\
\midrule

ADAM & treehill & 0.503 & 0.516 & 0.520 & 21.43 & 21.65 & 21.69 & 0.549 & 0.544 & 0.558 & \textbf{194} & \underline{3.47} \\
3DGS-LM \footnotemark[1] & treehill & 0.397 & 0.413 & 0.430 & 19.31 & 19.99 & 20.52 & 0.645 & 0.634 & 0.617 & 2648 & 42.58\\
ADAM-TR (Ours) & treehill & \underline{0.510} & \underline{0.524} & \underline{0.529} & \underline{21.47} & \underline{21.71} & \underline{21.75} & \underline{0.538} & \underline{0.533} & \underline{0.552} & \underline{210} & \textbf{3.46}\\
3DGS$^2$-TR (Ours) \footnotemark[2] & treehill & \textbf{0.520} & \textbf{0.530} & \textbf{0.533} & \textbf{21.73} & \textbf{21.84} & \textbf{21.84} & \textbf{0.533} & \textbf{0.529} & \textbf{0.543} & 600 & 4.26 \\
\midrule

ADAM & counter & 0.852 & 0.866 & 0.872 & 26.49 & 27.15 & 27.53 & 0.279 & 0.270 & 0.297 & \textbf{241} & \textbf{7.12} \\
3DGS-LM \footnotemark[1] & counter & 0.678 & 0.709 & 0.741 & 20.68 & 21.95 & 23.07 & 0.505 & 0.480 & 0.447 & 3531 & 61.56\\
ADAM-TR (Ours) & counter & \underline{0.859} & \underline{0.871} & \underline{0.876} & \underline{26.58} & \underline{27.25} & \underline{27.59} & \underline{0.272} & \underline{0.265} & \underline{0.290} & \underline{270} & \underline{7.24}\\
3DGS$^2$-TR (Ours) \footnotemark[2] & counter & \textbf{0.865} & \textbf{0.874} & \textbf{0.877} & \textbf{26.91} & \textbf{27.51} & \textbf{27.77} & \textbf{0.268} & \textbf{0.263} & \textbf{0.282} & 761 & 8.18 \\
\midrule

ADAM & stump & 0.508 & 0.517 & 0.520 & 22.77 & 22.83 & 22.86 & 0.542 & 0.538 & 0.550 & \textbf{182} & \textbf{2.60} \\
3DGS-LM \footnotemark[1] & stump & 0.384 & 0.397 & 0.416 & 20.44 & 20.85 & 21.30 & 0.670 & 0.662 & 0.643 & 2121 & 33.61\\
ADAM-TR (Ours) & stump & \underline{0.512} & \underline{0.528} & \underline{0.532} & \underline{22.78} & \underline{22.93} & \underline{22.99} & \underline{0.529} & \underline{0.524} & \underline{0.546} & \underline{187} & \underline{2.64}\\
3DGS$^2$-TR (Ours) \footnotemark[2] & stump & \textbf{0.525} & \textbf{0.534} & \textbf{0.537} & \textbf{22.98} & \textbf{23.08} & \textbf{23.13} & \textbf{0.522} & \textbf{0.518} & \textbf{0.532} & 589 & 3.39 \\
\midrule

ADAM & bonsai & 0.893 & 0.904 & 0.908 & 28.37 & 29.24 & 29.69 & 0.285 & 0.279 & 0.299 & \textbf{237} & \textbf{8.49} \\
3DGS-LM \footnotemark[1] & bonsai & 0.679 & 0.721 & 0.760 & 21.04 & 22.68 & 24.06 & 0.505 & 0.482 & 0.450 & 1921 & 43.60\\
ADAM-TR (Ours) & bonsai & \underline{0.901} & \underline{0.911} & \underline{0.914} & \underline{28.75} & \underline{29.67} & \underline{30.03} & \underline{0.273} & \underline{0.268} & \underline{0.288} & \underline{278} & \underline{8.63}\\
3DGS$^2$-TR (Ours) \footnotemark[2] & bonsai & \textbf{0.905} & \textbf{0.913} & \textbf{0.916} & \textbf{29.11} & \textbf{29.92} & \textbf{30.31} & \textbf{0.270} & \textbf{0.265} & \textbf{0.283} & 750 & 9.59 \\
\midrule

ADAM & bicycle & 0.480 & 0.498 & 0.504 & 21.58 & 21.84 & 21.95 & 0.523 & 0.516 & 0.539 & \textbf{185} & \textbf{3.70} \\
3DGS-LM \footnotemark[1] & bicycle & 0.346 & 0.360 & 0.376 & 18.96 & 19.47 & 19.89 & 0.676 & 0.666 & 0.652 & 2265 & 42.61\\
ADAM-TR (Ours) & bicycle & \underline{0.494} & \underline{0.516} & \underline{0.523} & \underline{21.70} & \underline{22.04} & \underline{22.11} & \underline{0.505} & \underline{0.496} & \underline{0.530} & \underline{203} & \underline{3.76}\\
3DGS$^2$-TR (Ours) \footnotemark[2] & bicycle & \textbf{0.511} & \textbf{0.526} & \textbf{0.532} & \textbf{21.92} & \textbf{22.17} & \textbf{22.24} & \textbf{0.494} & \textbf{0.488} & \textbf{0.510} & 565 & 4.53 \\
\midrule

ADAM & kitchen & 0.880 & 0.891 & 0.898 & 27.95 & 28.64 & 29.11 & 0.193 & 0.185 & 0.209 & \textbf{271} & \underline{8.41} \\
3DGS-LM \footnotemark[1] & kitchen & 0.592 & 0.640 & 0.698 & 20.94 & 22.38 & 23.66 & 0.490 & 0.450 & 0.401 & 3257 & 66.58\\
ADAM-TR (Ours) & kitchen & \underline{0.888} & \underline{0.897} & \underline{0.902} & \underline{28.03} & \underline{28.81} & \underline{29.20} & \underline{0.186} & \underline{0.179} & \underline{0.200} & \underline{307} & \textbf{8.35}\\
3DGS$^2$-TR (Ours) \footnotemark[2] & kitchen & \textbf{0.892} & \textbf{0.899} & \textbf{0.903} & \textbf{28.51} & \textbf{29.13} & \textbf{29.47} & \textbf{0.183} & \textbf{0.177} & \textbf{0.194} & 761 & 9.51 \\
\midrule

ADAM & flowers & 0.353 & 0.362 & 0.365 & 18.77 & 18.88 & 18.89 & 0.605 & 0.602 & 0.614 & \textbf{187} & \underline{3.41} \\
3DGS-LM \footnotemark[1] & flowers & 0.265 & 0.277 & 0.292 & 17.02 & 17.51 & 17.89 & 0.725 & 0.711 & 0.690 & 2654 & 48.45\\
ADAM-TR (Ours) & flowers & \underline{0.363} & \underline{0.374} & \underline{0.377} & \underline{18.81} & \underline{18.94} & \underline{18.97} & \underline{0.594} & \underline{0.590} & \underline{0.605} & \underline{202} & \textbf{3.39}\\
3DGS$^2$-TR (Ours) \footnotemark[2] & flowers & \textbf{0.370} & \textbf{0.377} & \textbf{0.379} & \textbf{18.93} & \textbf{19.02} & \textbf{19.04} & \textbf{0.592} & \textbf{0.589} & \textbf{0.599} & 555 & 4.18 \\
\midrule

ADAM & room & 0.870 & 0.879 & 0.883 & \underline{28.80} & 29.33 & 29.67 & 0.309 & 0.302 & 0.323 & \textbf{212} & \textbf{8.82} \\
3DGS-LM \footnotemark[1] & room & 0.718 & 0.766 & 0.797 & 21.96 & 23.79 & 25.32 & 0.488 & 0.467 & 0.438 & 2696 & 49.79\\
ADAM-TR (Ours) & room & \underline{0.877} & \underline{0.886} & \underline{0.890} & 28.77 & \underline{29.49} & \underline{29.81} & \underline{0.296} & \underline{0.290} & \underline{0.312} & \underline{238} & \underline{8.91}\\
3DGS$^2$-TR (Ours) \footnotemark[2] & room & \textbf{0.881} & \textbf{0.888} & \textbf{0.891} & \textbf{29.00} & \textbf{29.70} & \textbf{30.18} & \textbf{0.292} & \textbf{0.286} & \textbf{0.304} & 691 & 9.72 \\
\midrule

ADAM & garden & 0.660 & 0.678 & 0.687 & 23.87 & 24.11 & 24.29 & 0.370 & 0.363 & 0.387 & \textbf{202} & \textbf{4.36} \\
3DGS-LM \footnotemark[1] & garden & 0.373 & 0.394 & 0.428 & 19.88 & 20.53 & 21.23 & 0.674 & 0.644 & 0.595 & 2327 & 40.97\\
ADAM-TR (Ours) & garden & \underline{0.664} & \underline{0.685} & \underline{0.692} & \underline{23.98} & \underline{24.32} & \underline{24.45} & \underline{0.356} & \underline{0.349} & \underline{0.379} & \underline{217} & \underline{4.42}\\
3DGS$^2$-TR (Ours) \footnotemark[2] & garden & \textbf{0.672} & \textbf{0.688} & \textbf{0.694} & \textbf{24.12} & \textbf{24.37} & \textbf{24.50} & \textbf{0.353} & \textbf{0.346} & \textbf{0.369} & 591 & 5.26 \\
\midrule
\bottomrule
    \end{tabular}
    \begin{tablenotes}
    \item[1] 3DGS-LM results are reported @35, @75, and @150 iterations.
    \item[2] System implementation is not fully optimized. See Section~\ref{sec:limitations} for details.
    \end{tablenotes}
    \end{threeparttable}
    }
    \end{table*}

\clearpage
\section{Parameter-wise trust-region bounds based on squared Hellinger distance}
\label{sec:derivations}

We derive parameter-wise trust-region bounds such that when one parameter is updated while all other parameters fixed, the normalized distance between the Gaussian primitive before and after the update guarantees
\begin{equation*}
    \frac{H^2(G,G')}{\det(S)} \;<\; \epsilon.
\end{equation*}

\paragraph{Bound on the mean $\mu$.}

Let $\mu=(\mu_x,\mu_y,\mu_z), \mu'=(\mu_x',\mu_y',\mu_z'), \Delta\mu = \mu' - \mu$. We first let $\Delta\mu_x \neq 0, \Delta\mu_y = 0, \Delta\mu_z = 0$, then 
\begin{equation*}
    H^2(G,G') = C \left(1 - \exp\left(-\frac{\Delta\mu^T \Sigma^{-1} \Delta\mu}{8} \right)\right)  = C\left(1 - \exp\left(-\frac{\Delta\mu_x^2}{8\Sigma_{xx}}  \right)\right).
\end{equation*}
Setting $H^2(G, G') / \det(S) < \epsilon$ gives $\alpha \left(1 - \exp\left(-\frac{\Delta\mu_x^2}{8\Sigma_{xx}}  \right)\right) < \epsilon$.
After rearranging the terms, we have
\begin{align*}
    |\Delta\mu_x|  < \sqrt{-8\,\Sigma_{xx}\,\ln\!\left(1 - \frac{\epsilon}{\alpha}\right)}.
\end{align*}
Using similar derivation, we have
\begin{align*}
   |\Delta\mu_y| &< \sqrt{-8\,\Sigma_{yy}\,\ln\!\left(1 - \frac{\epsilon}{\alpha }\right)}, \\
   |\Delta\mu_z| &< \sqrt{-8\,\Sigma_{zz}\,\ln\!\left(1 - \frac{\epsilon}{\alpha }\right)} .
\end{align*}

\paragraph{Bound on the scale matrix $S$.}
Let
\begin{equation*}
  S=\begin{bmatrix} S_x && \\ & S_y & \\ && S_z \end{bmatrix},
\qquad
S'=\begin{bmatrix} S_x' && \\ & S_y' & \\ && S_z' \end{bmatrix},  
\end{equation*}
and define $\Delta S=S'-S$ with  differences $\Delta S_x,\Delta S_y,\Delta S_z$. 

Note that $C = \alpha \mathcal{C} \det(\Sigma)^{1/2} = \alpha \mathcal{C} \det(S), C' = \alpha \mathcal{C} \det(S')$, we first
let $\Delta S_x \neq 0$ while $\Delta S_y = 0,\Delta S_z = 0$, and denote $\rho_x = \alpha S_y S_z$.
Then 
\begin{equation*}
   H^2(G,G') = \frac12 (\rho_x S_x + \rho_x S_x') - \rho_x(S_x S_x')^{1/2} \frac{(S_x)^{\frac12} (S_x')^{\frac12}S_y S_z}{(\frac{S_x^2 + S_x'^2}{2})^{\frac{1}{2}} S_y S_z} \\ = \rho_x \frac{S_x + S_x'}{2} - \rho_x \frac{S_x S_x'}{\left(\frac{S_x^2 + S_x'^2}{2}\right)^{\frac{1}{2}}}  
\end{equation*}

Since $S_x' = S_x + \Delta S_x$, we have 
$\frac{\mathrm{d}}{\mathrm{d}\Delta S_x}H^2 = \rho_x \left(\frac12 + \frac{-2S_x^2}{(2S_x^2+\Delta S_x)^2}\right)$, and
$\frac{\mathrm{d}^2}{\mathrm{d}\Delta S_x^2}H^2 = \rho_x (4S_x^2 \cdot (2S_x+\Delta S_x)^{-3})$. We thus can approximate $H^2(G, G') \approx \frac{\rho_x}{2S_x}\Delta S_x^2$.

Setting $H^2(G,G') / \det{S} < \epsilon$ gives 
\begin{equation*}
    |\Delta S_x| < \sqrt{\frac{2S_x \det(S) \epsilon}{\rho_x}} = \sqrt{\frac{2S_x^2  \epsilon}{\alpha }}.
\end{equation*}
We can similarly obtain
\begin{align*}
   |\Delta S_y| < \sqrt{\frac{2 S_y^2\,\epsilon}{\alpha }}, \;\text{and}\;
   |\Delta S_z| < \sqrt{\frac{2 S_z^2\,\epsilon}{\alpha }}.
\end{align*}

\paragraph{Bound on the opacity $\alpha$.}

Let $\rho_\alpha = \det(\Sigma)^{1/2} = \det(S)$, then
\begin{equation*}
    H^2(G,G') = \rho_\alpha \left(\frac{\alpha + \alpha'}{2} - (\alpha \alpha')^{1/2} \right).
\end{equation*}

Let $\alpha' = \alpha + \Delta \alpha$, then 
\begin{equation*}
    \frac{\mathrm{d}}{ \mathrm{d} \Delta \alpha} H^2 = \rho_\alpha\left(\frac12 - \frac12 \alpha (\alpha^2 + \alpha \Delta \alpha )^{-1/2}\right)
\end{equation*}
and
\begin{equation*}
   \frac{\mathrm{d}^2}{\mathrm{d} \Delta \alpha^2} H^2 = \rho_\alpha\left(\frac14 \alpha^2 (\alpha^2 + \alpha \Delta \alpha )^{-3/2}\right) .
\end{equation*}

Thus, we can approximate 
\begin{equation*}
    H^2(G,G') \approx \frac{\rho_\alpha}{4\alpha}\Delta\alpha^2.
\end{equation*}

Setting $H^2(P, Q) / \det(S) < \epsilon$ gives 

\begin{equation*}
    |\Delta\alpha| < \sqrt{\frac{4\alpha \det(S) \epsilon}{\rho_\alpha}} = \sqrt{4\alpha \epsilon}. 
\end{equation*}

\paragraph{Bound on the color $\mathcal{C}$.}

Let $\mathcal{C}=\mathcal{C}_r+\mathcal{C}_g+\mathcal{C}_b$ and similarly define $\mathcal{C}'$ and $\Delta\mathcal{C}_c=\mathcal{C}_c'-\mathcal{C}_c$ for $c\in\{r,g,b\}$. First, let $\Delta\mathcal{C}_r\neq 0$ and $\Delta\mathcal{C}_g=\Delta\mathcal{C}_b=0$.
Let $\rho_r = \alpha \det(\Sigma)^{1/2} = \alpha \det(S)$, then 
\begin{equation*}
    H^2(G,G') = \rho_r \left(\frac12 (\mathcal{C}_r + \mathcal{C}_r') - (\mathcal{C}_r \mathcal{C}_r')^{1/2} \right). 
\end{equation*}

We can approximate 
\begin{equation*}
    H^2(G, G') \approx \frac{\rho_r}{4 \mathcal{C}_r} \Delta \mathcal{C}_r^2 .
\end{equation*}

Setting $H^2(P, Q) / \det(S) < \epsilon$ gives 
\begin{equation*}
    |\Delta \mathcal{C}_r| < \sqrt{\frac{4 \mathcal{C}_r \det(S) \epsilon}{\rho_r}} = \sqrt{\frac{4 \mathcal{C}_r \epsilon}{\alpha}}.
\end{equation*}
Using similar derivation, we also have
\begin{align*}
   |\Delta\mathcal{C}_g| < \sqrt{\frac{4\mathcal{C}_g\,\epsilon}{\alpha}}, \;
   |\Delta\mathcal{C}_b| < \sqrt{\frac{4\mathcal{C}_b\,\epsilon}{\alpha}} .
\end{align*}

\paragraph{Bound on the rotation $R$.}
The rotation matrix $R$ is parameterized by a unnormalized quaternion $\tilde{q}=(q_x,q_y,q_z,q_w)$ with squared norm
$r^2=\|q\|^2 = q_x^2 + q_y^2 + q_z^2 + q_w^2$, yielding
\begin{equation*}
    R =
\begin{bmatrix}
1 - \frac{2(q_y^2 + q_z^2)}{r^2} & \frac{2(q_x q_y - q_w q_z)}{r^2} & \frac{2(q_x q_z + q_w q_y)}{r^2} \\
\frac{2(q_x q_y + q_w q_z)}{r^2} & 1 - \frac{2(q_z^2 + q_x^2)}{r^2} & \frac{2(q_y q_z - q_w q_x)}{r^2} \\
\frac{2(q_x q_z - q_w q_y)}{r^2} & \frac{2(q_y q_z + q_w q_x)}{r^2} & 1 - \frac{2(q_x^2 + q_y^2)}{r^2}
\end{bmatrix}.
\end{equation*}
We can similarly construct $R'$, which is parameterized by another unnormalized quaternion $q'=(q_x',q_y',q_z',q_w')$ with squared norm
$r'^2=\|q'\|^2 = q_x'^2 + q_y'^2 + q_z'^2 + q_w'^2$.

Since $\Sigma = R^TS^TSR, \Sigma' = R'^T S^T S R'$
and $R, R'$ are orthogonal, we have $\det(\Sigma) = \det(\Sigma') = \det(S)^2$. Let $\Delta R = R^{-1}R'$ also be an orthogonal rotation matrix.
Then 
\begin{equation*}
    H^2(G,G') = C\left(1 - \frac{\det(S)}{\det(\frac{\Sigma + \Sigma'}{2})^{\frac12}}\right) = C\left(1 - \frac{\det(S)}{\det(\frac{S^2 + \Delta R^T S^2 \Delta R}{2})^{\frac12}}\right).
\end{equation*}

Setting $H^2(G,G') / \det(S) < \epsilon$ and rearranging the terms gives
\begin{align*}
  1 - \frac{\det(S)}{\det(\frac{S^2 + \Delta R^T S^2 \Delta R}{2})^{\frac12}} < \frac{\epsilon}{\alpha }. 
\end{align*}
To satisfy the above inequality, it is sufficient that $\det\left(\frac{S^2 + \Delta R^T S^2 \Delta R}{2}\right) < \left(\frac{\det(S)}{1 - \epsilon / \alpha}\right)^2$, which is sufficient if $\det\left(S^2 \left(\frac{I + S^{-2}\Delta R^T S^2 \Delta R}{2}\right)\right) < \left(\frac{\det(S)}{1 - \epsilon / \alpha }\right)^2$. Since for two matrices $A$ and $B$, $\det(AB)=\det(A)\det(B)$, we only need $\det\left(\frac{I + S^{-2}\Delta R^T S^2 \Delta R}{2}\right) < \left(\frac{1}{1 - \epsilon / \alpha}\right)^2$, equivalently,
\begin{equation*}
    \det\left(I + \left(-I + \frac{I + S^{-2}\Delta R^T S^2 \Delta R}{2}\right)\right) < \left(\frac{1}{1 - \epsilon / \alpha}\right)^2.
\end{equation*}
By the relation $\det (I+X)\leq \exp(\text{tr}(X))$, it is sufficient to have
\begin{equation*}
    \text{tr}\left(\frac{I + S^{-2}\Delta R^T S^2 \Delta R}{2} - I\right) < -2 \ln \left(1 - \frac{\epsilon}{\alpha }\right).
\end{equation*}
By the linearity of trace, we only need to show
\begin{equation}\label{appendix_eq1}
    \text{tr}(S^{-2}\Delta R^T S^2 \Delta R) - 3 < -4 \ln\left(1 - \frac{\epsilon}{\alpha}\right).
\end{equation}

Next, we investigate the approximation of $ \text{tr}(S^{-2}\Delta R^T S^2 \Delta R)$. For an unnormalized quaternion $ \tilde{q} = (q_x,q_y,q_z,q_w)$ with $r^2=\| \tilde{q} \|^2$, the unnormalized rotation matrix is
\begin{equation*}
    \tilde{R}(\tilde{q}) = \begin{bmatrix} r^2 - 2(q_y^2 + q_z^2) & (2q_xq_y - 2q_wq_z) & (2q_xq_z + 2q_wq_y) \\ (2q_xq_y + 2q_wq_z) & r^2 - 2(q_z^2 + q_x^2) & (2q_yq_z - 2q_wq_x) \\ (2q_xq_z - 2q_wq_y) & (2q_yq_z + 2q_wq_x) & r^2 - 2(q_x^2 + q_y^2) \end{bmatrix}
\end{equation*}
which can be normalized as $R = \tilde{R} / r^2$. 

Let the update to the unnormalized quaternion be $\Delta q = (\Delta q_x, \Delta q_y, \Delta q_z, \Delta q_w)$. We consider an update in the direction of \(\Delta q\) with a step size \(a\), denoted as $ q' = \tilde q + a \Delta q$ with $r'^2 = (q_x + a\Delta q_x)^2 + (q_y + a\Delta  q_y)^2 + (q_z + a\Delta q_z)^2 + (q_w + a\Delta q_w)^2  $

Let \(R' = \tilde R(q') / r'^2 = R\Delta R = R(I+E)\), where \(R, \Delta R, R'\) are orthogonal, and
\(E = R^TR' - I = R^T \tilde R'/r'^2 - I\).
Then with the relation $\text{tr}(S^{-2}\Delta R^T S^2 \Delta R) = \text{tr}(S^{-1} \Delta R^T S S \Delta R S^{-1})  = \|S \Delta R S^{-1}\|_F^2  = \|S (I + E) S^{-1}\|_F^2$, we have
\begin{align*}
    \frac{\partial}{\partial x} \|S(I+E)S^{-1}\|_F^2 & = 2\text{tr} ((S(I + E)S^{-1})^T (S \partial_x E S^{-1})),  \\
    \frac{\partial^2}{\partial x^2} \|S(I+E)S^{-1}\|_F^2 & = 2 \| S \partial_x E S^{-1} \|_F^2 + 2 \text{tr} ((S(I + E)S^{-1})^T (S \partial_x^2E S^{-1})), 
\end{align*}
where the partial derivative to $x$ is short for the partial derivative to $q_x$. We have similar formulas for the partial derivative to $q_y,q_z,q_w$. For the variable $q_c \in \{q_x, q_y, q_z, q_w\}$, it can be computed that $\partial_c E = R^T (r^{-2}\partial_c \tilde{R} - 2q_c r^{-4}\tilde{R})$, and 
\begin{align*}
  \partial_c^2E &= R^T (-2q_cr^{-4}\partial_c\tilde{R} + r^{-2} \partial_c^2\tilde{R} +6 q_c^2 r^{-6} \tilde{R}_c - 2q_c q^{-4} \partial_c \tilde{R} - 2 q^{-4} \tilde{R} + 2q_c^2 r^{-6} \tilde{R}) \\ & = R^T (-2q_cr^{-4}\partial_c\tilde{R} + r^{-2} \partial_c^2\tilde{R} +8 q_c^2 r^{-6} \tilde{R} - 2 q_c r^{-4} \partial_c\tilde{R} - 2 r^{-4} \tilde{R}) . 
\end{align*}
Denote \(T(\Delta q) = \|S(I+E)S^{-1}\|_F^2\), then when $\Delta q_x\neq 0,\Delta q_y=0,\Delta q_z=0,\Delta q_w=0$, we have \(T(\Delta q_x) = 3, T'(\Delta q_x) = 0, \beta_x = T''(\Delta q_x) = 2 \| S \partial_x E S^{-1} \|_F^2 + 2 \text{tr} (\partial_x^2E)\).
We thus can approximate $ \text{tr}(S^{-2}\Delta R^T S^2 \Delta R)$ by its second-order Taylor expansion and obtain
the approximation of \eqref{appendix_eq1} as $3 + \frac12 \beta_x (\Delta q_x)^2 - 3 < -4 \ln (1 - \epsilon / \alpha )$. Rearranging the terms, we obtain 
\begin{equation*}
    |\Delta q_x| < \sqrt{-8 / \beta_x \ln (1 - \epsilon / \alpha)}.
\end{equation*}
Using similar derivation, we also obtain the bounds
\begin{equation*}
    |\Delta q_y| < \sqrt{-8 / \beta_y \ln (1 - \epsilon / \alpha)},\; 
    |\Delta q_z| < \sqrt{-8 / \beta_z \ln (1 - \epsilon / \alpha)},\text{ and } 
    |\Delta q_w| < \sqrt{-8 / \beta_w \ln (1 - \epsilon / \alpha)}.
\end{equation*}

\end{document}